\documentclass[10pt,twocolumn,letterpaper]{article}

\usepackage[hyphens]{url}
\usepackage{enumitem}
\usepackage{subcaption}
\usepackage{array}
\usepackage{booktabs}
\usepackage{adjustbox}
\usepackage{multirow}
\usepackage[inkscapelatex=false]{svg}
\definecolor{cvprblue}{rgb}{0.21,0.49,0.74}
\usepackage{caption}
\usepackage{enumitem}
\setitemize{noitemsep,topsep=0pt,parsep=0pt,partopsep=0pt}
\usepackage{pgfplots}
\usepackage{capt-of}
\usepackage{float}
\usepackage{authblk} 
\usepackage[pagebackref,breaklinks,colorlinks,allcolors=cvprblue]{hyperref}
\usepackage[accsupp]{axessibility}
\usepackage[pagenumbers]{cvpr}

\newcommand{\MethodLong}{Diffusion Fourier Neural Opeartor\xspace}
\newcommand{\MethodShort}{DiffFNO\xspace}
\newcommand{\NewFNOLong}{Weighted Fourier Neural Operator\xspace}
\newcommand{\NewFNOShort}{WFNO\xspace}
\newcommand{\NewFNOFeatureLong}{Mode Rebalancing\xspace}
\newcommand{\NewFNOFeatureShort}{MR\xspace}
\newcommand{\NewGatedFusionLong}{Gated Fusion Mechanism\xspace}
\newcommand{\NewGatedFusionShort}{GFM\xspace}
\newcommand{\NewSolverLong}{Adaptive Time-Step\xspace}
\newcommand{\NewSolverShort}{ATS\xspace}
\newcommand{\SpatialNOLong}{Attention-based Neural Operator\xspace}
\newcommand{\SpatialNOShort}{AttnNO\xspace}

\pgfplotsset{compat=1.18} 

\def\confName{CVPR}
\def\confYear{2025}

\title{DiffFNO: Diffusion Fourier Neural Operator}

\author{
\begin{tabular}{ccc}
Xiaoyi Liu$^{1, *}$ \quad Hao Tang$^{2,\dagger}$\\
$^{1}$Washington University in St. Louis \quad $^{2}$Peking University \\
{\tt\small jasonl@wustl.edu} \quad {\tt\small haotang@pku.edu.cn}
\end{tabular}
}

\begin{document}

\twocolumn[{
\maketitle
\begin{center}
    \captionsetup{type=figure}
    \begin{minipage}{0.48\textwidth}
        \subcaptionbox{\textbf{PSNR and inference time for $\times$4 super-resolution}}[0.9\linewidth]{
        \begin{tikzpicture}
            \begin{axis}[
                width=8cm, height=6cm, 
                xlabel={Inference Time (ms)},
                ylabel={PSNR (dB)},
                xlabel near ticks, 
                ylabel near ticks, 
                ymin=28, ymax=31, 
                xmin=130, xmax=265, 
                enlargelimits=0.05, 
                grid=both,
                scaled ticks=false,
                clip=false,
            ]
    
            \addplot[red, mark=star, mark size=4pt, mark options={solid, line width=2pt}] coordinates {(141, 30.88)};
            \node[right] at (axis cs:141, 30.88) {\textbf{ DiffFNO (Ours)}};
    
            \addplot[blue, only marks, mark=*] coordinates {(225, 29.31)};
            \node[above] at (axis cs:225, 29.31) {LIIF \cite{SR-LIIF}};
    
            \addplot[blue, only marks, mark=*] coordinates {(210, 28.79)};
            \node[above] at (axis cs:210, 28.79) {LTE \cite{SR-LTE}};
    
            \addplot[blue, only marks, mark=*] coordinates {(240, 29.70)};
            \node[above] at (axis cs:240, 29.70) {LIT \cite{SR-LIT}};
    
            \addplot[blue, only marks, mark=*] coordinates {(232, 30.40)};
            \node[below] at (axis cs:232, 30.40) {LMI \cite{SR-LMI}};
    
            \addplot[blue, only marks, mark=*] coordinates {(248, 30.46)};
            \node[above] at (axis cs:248, 30.46) {HiNOTE \cite{SR-HiNOTE}};
    
            \addplot[blue, only marks, mark=*] coordinates {(195, 28.23)};
            \node[above] at (axis cs:195, 28.23) {Meta-SR \cite{SR-Meta-SR}};
    
            \addplot[blue, only marks, mark=*] coordinates {(147, 30.05)};
            \node[above] at (axis cs:147, 30.05) {SRNO \cite{SR-SRNO}};
            \end{axis}
        \end{tikzpicture}
        }
    \end{minipage}
    \hfill
    \begin{minipage}{0.48\textwidth}
        \subcaptionbox{\textbf{\MethodShort vs SRNO \cite{SR-SRNO}}}[0.9\linewidth]{
            \includegraphics[width=\linewidth]{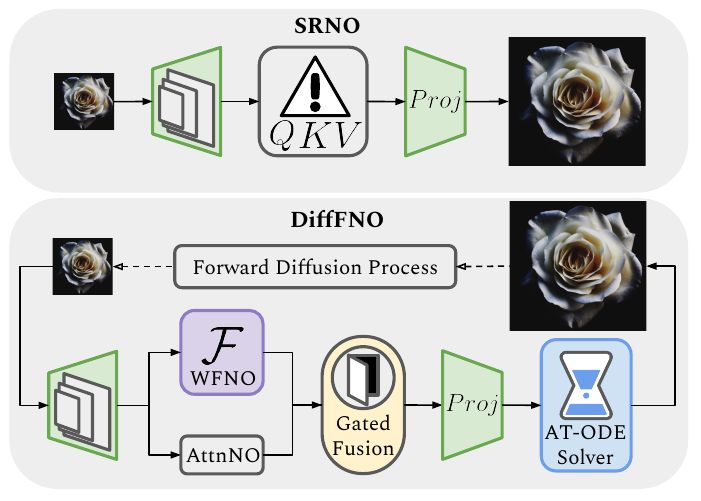}
        }
    \end{minipage}
    \caption{(a) All models use the EDSR-baseline \cite{EDSR} encoder, except HiNOTE \cite{SR-HiNOTE} which has its own. (b) Compared to SRNO \cite{SR-SRNO}, \MethodShort is strengthened by the fusion of spectral and spatial features and efficient refinement by a diffusion process.}
    \label{fig:teaser}
\end{center}
}]

\renewcommand{\thefootnote}{} 
\footnotetext{$^\dagger$Corresponding author.\\ \indent \indent $^*$Work Done during the visit at Peking University.}
\renewcommand{\thefootnote}{\arabic{footnote}} 

\begin{abstract}
We introduce \MethodShort, a novel diffusion framework for arbitrary-scale super-resolution strengthened by a \NewFNOLong (\NewFNOShort). 
\NewFNOFeatureLong in \NewFNOShort effectively captures critical frequency components, significantly improving the reconstruction of high-frequency image details that are crucial for super-resolution tasks. 
\NewGatedFusionLong (\NewGatedFusionShort) adaptively complements \NewFNOShort's spectral features with spatial features from an \SpatialNOLong (\SpatialNOShort). This enhances the network's capability to capture both global structures and local details. 
\NewSolverLong (\NewSolverShort) ODE solver, a deterministic sampling strategy, accelerates inference without sacrificing output quality by dynamically adjusting integration step sizes \NewSolverShort. 
Extensive experiments demonstrate that \MethodShort achieves state-of-the-art (SOTA) results, outperforming existing methods across various scaling factors by a margin of \textbf{2–4 dB in PSNR}, including those beyond the training distribution. It also achieves this at lower inference time (Fig. \ref{fig:teaser} (a)). Our approach sets a new standard in super-resolution, delivering both superior accuracy and computational efficiency.
\end{abstract}    
\section{Introduction}
\label{sec:intro}
Image super-resolution (SR) reconstructs high-resolution (HR) images from low-resolution (LR) inputs, recovering lost fine details to enhance visual quality. SR is crucial for applications like medical imaging \cite{MedicalImageSR}, satellite imagery \cite{SatelliteSR, RemoteSensingSR}, and video games \cite{VideoGameSR}. The challenge in SR lies in its ill-posed nature: multiple HR images can correspond to the same LR input due to information loss during downsampling. This ambiguity requires sophisticated algorithms capable of inferring plausible and perceptually accurate high-frequency content from limited data.

Deep learning, particularly Convolutional Neural Networks (CNNs) \cite{RDN}, has significantly advanced SR. Dong et al. introduced SRCNN \cite{SRCNN}, demonstrating the effectiveness of end-to-end learning for SR. Subsequent models achieved remarkable performance using deeper architectures and attention mechanisms \cite{EDSR, RCAN, TransformerSR-1, TransformerSR-2, SwinIR}.

\begin{figure*}[t]  
  \centering
 \includegraphics[width=\textwidth,keepaspectratio]{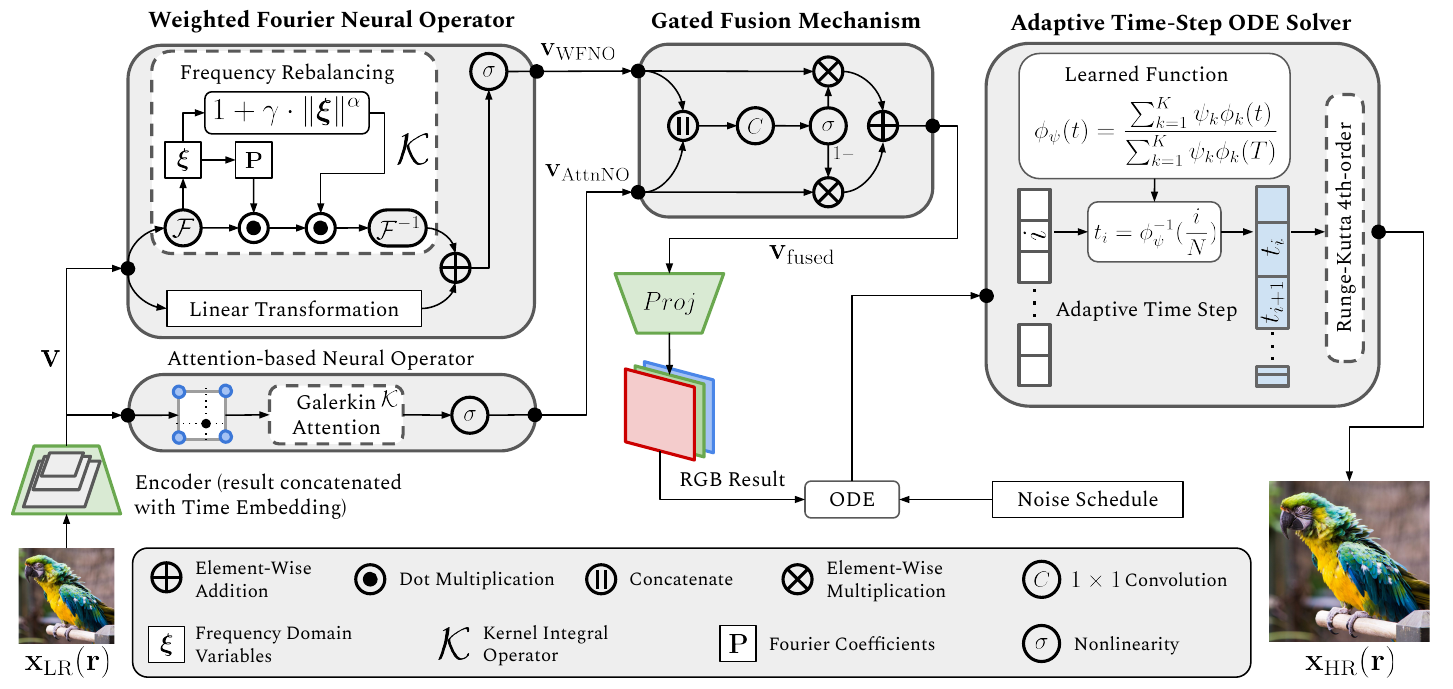}  
    \caption{The proposed \MethodLong (\MethodShort) architecture for arbitrary-scale super-resolution begins by lifting a low-resolution input image $\mathbf{x}_\text{LR}(\mathbf{r})$ into a feature space using a convolutional encoder. Features extracted by the \NewFNOLong (\NewFNOShort) and an \SpatialNOLong (\SpatialNOShort) are combined using a \NewGatedFusionLong (\NewGatedFusionShort). The fused features are then projected into RGB space, where \NewSolverLong (\NewSolverShort) ODE solver efficiently completes the reverse diffusion process with both accuracy and speed. This pipeline generates $\mathbf{x}_\text{HR}(\mathbf{r})$,  a high-resolution version of the input image.}
    \label{fig:short-a}
    \vspace{-0.4cm}
\end{figure*}

Diffusion models have emerged as powerful generative frameworks modeling complex data distributions via iterative denoising processes \cite{DDPM, Diffusion-Implicit, Diffusion-Implicit-2023}. Their ability to generate high-fidelity images is well-suited for inferring missing fine details. In SR, diffusion models progressively refine an LR image by modeling the conditional distribution of HR images given the LR input \cite{CascadedDiffusionSR, Diffusion-SR, Diffusion-SR-Srdiff, Diffusion-SR-prior}. This iterative process reconstructs intricate textures and high-frequency components, producing realistic outputs.

However, diffusion models are computationally intensive due to the iterative reverse diffusion process \cite{SDE}. To address this, recent research explores efficient sampling strategies to accelerate reverse diffusion. One approach is approximating the diffusion process through deterministic Ordinary Differential Equation (ODE), which can be solved in fewer steps \cite{DPM-Solver}. This accelerates inference and provides consistent, reproducible results.

Arbitrary-scale SR models \cite{SR-Meta-SR, SR-LTE, SR-LIIF}, which can upsample images at user-defined scales beyond those seen in training, have gained attention in recent years. Methods involving attention mechanisms \cite{SR-LIT} and representing images as continuous functions \cite{SR-LMI} have been explored. Operator-learning methods such as Super-Resolution Neural Operators (SRNO) \cite{SR-SRNO} and HiNOTE \cite{SR-HiNOTE} have further advanced this field. However, the inherent differences between physics simulations and real-world images introduce challenges from computational demands to the difficulty in restoring high-frequency details.

\noindent \textbf{To address these limitations, our contributions are}:

\noindent (1) We propose \NewFNOLong (\NewFNOShort) strengthened by iterative refinement from a diffusion framework for high-frequency reconstruction, detailed in Fig.~\ref{fig:short-a}. Through \NewFNOFeatureLong (\NewFNOFeatureShort), \NewFNOShort learns to emphasize the most critical frequency components. This greatly enhances high-frequency image detail reconstruction, overcoming the limitations of standard FNOs and MLPs, which underrepresent such details due to mode truncation and spectral bias, respectively. 
(2) We develop \NewGatedFusionLong (\NewGatedFusionShort) to dynamically adjust the influence of Fourier space features from \NewFNOShort and complementary spatial domain features from an \SpatialNOLong (\SpatialNOShort). \SpatialNOShort is lightweight, sharing an encoder with and running in parallel to \NewFNOShort.
(3) Additionally, we present \NewSolverLong (\NewSolverShort) ODE solver, which flexibly adjusts integration step sizes based on data characteristics by assessing the complexity of image regions, thereby reducing computational overhead without compromising quality.
(4) \MethodShort achieves state-of-the-art results on multiple SR benchmarks, outperforming existing methods by \textbf{2–4 dB in PSNR} in reconstruction quality. It also offers competitive inference time as Fig. \ref{fig:teaser} (a) shows. \MethodShort remains robust across various upscaling factors—even those unseen during training.
\section{Related Work}
\label{sec:formatting}

\noindent\textbf{Neural Operators and Fourier Methods.}
\textit{Neural Operators (NO)} \cite{NO} have emerged as a powerful framework for learning mappings between infinite-dimensional function spaces, providing resolution-invariant models that generalize across different input resolutions. Unlike traditional neural networks that map finite-dimensional vectors to other vectors, neural operators learn mappings from functions to functions \cite{FNO-Geometry}, making them well-suited for tasks involving continuous data or data at varying resolutions.

\textit{Multi-Layer Perceptrons (MLPs)} often exhibit a spectral bias, favoring low-frequency functions \cite{MLP-Spectral-Bias}. This limits their ability to capture fine textures and sharp edges. To overcome these limitations, techniques like positional encodings and Fourier feature mappings capture high-frequency details by embedding input coordinates into a higher-dimensional sinusoidal space, allowing the network to represent complex patterns \cite{Fourier-Feature-Map, Implicit-Neural-Representation}.

\textit{Fourier Neural Operator (FNO)} \cite{FNO} is a variant of NO that uses spectral convolution to efficiently capture global data patterns, modeling long-range dependencies with lower computational complexity than traditional CNNs. In physics and climate settings \cite{PINO, FNO-Near-Surface, FNO-downscale}, FNOs can handle arbitrary input resolutions without retraining. Although successful, FNOs may still lose high-frequency information due to mode truncation (discarding higher-frequency Fourier modes). This loss impairs tasks such as SR that rely on detailed reconstruction \cite{Wavelet-NO, Factorized-FNO, Multi-Wavelet}.

The \NewFNOFeatureLong mechanism in \NewFNOShort overcomes these limitations. Instead of being truncated, all Fourier modes are preserved, with additional learnable weights to modulate their impact on reconstruction. Fine-grained feature representation is further enhanced by \SpatialNOShort, which captures local details by processing data directly in the spatial domain

\noindent\textbf{Diffusion-Based SR and Efficient Sampling.}
Diffusion models have gained prominence as powerful generative models capable of producing high-quality images through iterative denoising techniques \cite{DDPM, SR3}. In the context of SR, diffusion models have been employed to model the conditional distribution of HR images given LR inputs, achieving higher resolutions after progressive enhancement \cite{Diffusion-SR, Diffusion-SR-Latent}.

Despite their effectiveness, diffusion models are computationally intensive due to the large number of time steps required in the reverse diffusion process. Such computational demands pose significant challenges for practical applications \cite{Diffusion-Elucidate}, especially in real-time or resource-constrained settings. Current solutions include: 
(i) \textit{Deterministic Sampling via ODE Solvers}: By reformulating the stochastic reverse diffusion as a deterministic ODE, advanced ODE solvers can be employed to reduce the number of sampling steps \cite{Diffusion-On-Fast-Sampling}. Methods like Denoising Diffusion Implicit Models (DDIM) \cite{Diffusion-Implicit} and DPM-Solver \cite{DPM-Solver, Analytic-DPM} have demonstrated the ability to generate high-quality images with significantly fewer steps.
(ii) \textit{Operator Learning for Fast Sampling}: Neural operators accelerates sampling by learning the solution operator of the reverse diffusion process \cite{Diffusion-Operator-Learning, Langevin-Diffusion}.
(iii) \textit{Progressive Distillation}: Training a distilled model to approximate the behavior of the full diffusion model allows faster sampling with fewer steps \cite{Diffusion-Progressive-Distillation, Early-Exiting}. Although effective, this method may require extensive retraining and potentially compromise image quality for increased speed.

Applying these acceleration methods to diffusion-based SR enables faster inference while maintaining high image quality. With efficient sampling methods, diffusion models become more practical for SR tasks, balancing performance and computational efficiency \cite{Compositional}.

\MethodShort adopts (i) for its simplicity and the robustness of established numerical methods. We also strength it with the \NewSolverShort strategy, which adjusts integration step sizes adaptively to balance speed and quality.
\section{The Proposed \MethodShort}

\subsection{Network Architecture and Novel Components}
An overview of the proposed \MethodShort is shown in Fig. \ref{fig:teaser} (b). It has three parts: 
(i) A CNN encoder extracts features from LR images. Unlike the simple linear transformations in standard FNO setups for physics simulations, our encoder is tailored for SR, extracting complex patterns and textures needed for high-quality reconstructions. We use the EDSR-baseline \cite{EDSR} and RDN models \cite{RDN} in our experiments. 
(ii) \NewFNOShort and \NewGatedFusionShort: \NewFNOShort captures both global and local details alongside the \SpatialNOShort. \NewGatedFusionShort combines these into a unified HR feature map, which is then projected into RGB. 
(iii) \NewSolverShort ODE solver accelerates inference speed by taking fewer, larger, and dynamically adjusted steps toward the reconstructed HR image. 
Fig. \ref{fig:short-a} illustrates these components in detail.

The network minimizes the difference between the predicted image and the true image. The loss function is:
\begin{equation}
\mathcal{L}(\theta) = \mathbb{E}_{t, \mathbf{x}_0} \left[ \left\| s_\theta(\mathbf{x}_t, t) - \nabla_{\mathbf{x}_t} \log p_t(\mathbf{x}_t | \mathbf{x}_0) \right\|_2^2 \right],
\label{eq:loss_function}
\end{equation}
where $\mathbf{x}_t$ (i.e. $\mathbf{x}_\text{LR}$) is obtained by adding noise to $\mathbf{x}_0$ (i.e. $\mathbf{x}_\text{HR}$). $s_\theta(\mathbf{x}_t, t)$ is the neural network approximating the true score function. $\nabla{\mathbf{x}_t} \log p_t(\mathbf{x}_t | \mathbf{x}_0)$ is the true score function.

\subsection{\NewFNOLong}
The Fourier Neural Operator (FNO) \cite{FNO} is an efficient NO variant designed to learn mappings between function spaces. It operates directly on inputs of arbitrary resolutions, performing upscaling by mapping low-resolution inputs to high-resolution outputs. It first transforms the input data into the frequency domain, applies the learned filters, and then transforms the data back to the spatial domain. Spectral convolution and mode truncation greatly enhance computational efficiency.

Let $\mathbf{x}_{\text{LR}}(\mathbf{r})$ denote the low-resolution input function (e.g., an image), where $\mathbf{r} \in \mathbb{R}^2$ represents spatial coordinates. The goal is to learn an operator $ \mathcal{G}$ such that:
\begin{equation}
\mathbf{x}_{\text{HR}}(\mathbf{r}) = \mathcal{G}[\mathbf{x}_{\text{LR}}(\mathbf{r})],
\label{eq:operator_goal}
\end{equation}
where $\mathbf{x}_{\text{HR}}(\mathbf{r})$ is the output function (e.g., the super-resolved image).
The FNO models $ \mathcal{G}$ by stacking:
\begin{equation}
\mathbf{v}_{l+1}(\mathbf{r}) = \sigma\left( \mathcal{W}_l \mathbf{v}_l(\mathbf{r}) + \mathcal{K}_l \mathbf{v}_l(\mathbf{r}) \right),
\label{eq:fno_layer}
\end{equation}
where $\mathbf{v}_l(\mathbf{r})$ is the feature representation at layer $l$ evaluated at spatial location $\mathbf{r}$, and $\mathbf{v}_{l+1}(\mathbf{r})$ is its updated representation in the following layer;
$\sigma$  is a nonlinear activation function;
$\mathcal{W}_l$ is a linear transformation. $\mathcal{K}_l$, the integral operator at layer $l$, transforms the features into the Fourier domain. Fourier modes are then truncated for computational efficiency. Global convolution is performed with a pointwise multiplication between the transformed features and the learned Fourier coefficients. 
\begin{equation}
\mathcal{K}_l \mathbf{v}_l(\mathbf{r}) = \mathcal{F}^{-1} \left( \mathbf{P}_l(\boldsymbol{\xi}) \cdot \mathcal{F}[\mathbf{v}_l](\boldsymbol{\xi}) \right),
\label{eq:integral_operator}
\end{equation}
where $\mathcal{F}$ and $\mathcal{F}^{-1}$ denote the Fourier and inverse Fourier transforms, respectively. $\boldsymbol{\xi}$ is the frequency domain variables. $\mathcal{F}[\mathbf{v}_l](\boldsymbol{\xi})$ is the Fourier transform of $\mathbf{v}_l$, evaluated at frequency $\boldsymbol{\xi}$. $\mathbf{P}_l(\boldsymbol{\xi})$ is a complex-valued tensor of learnable parameters representing the Fourier domain filters.

However, mode truncation underrepresents high-frequency components that are critical to SR of real-world images. To address this limitation, we introduce \NewFNOFeatureLong. A learned weighting function $\mathbf{w}_l(\boldsymbol{\xi})$ is applied to the Fourier modes to amplify or attenuate specific frequency components. It is defined at layer $l$ as:
\begin{equation}
\mathbf{w}_l(\boldsymbol{\xi}) = 1 + \gamma_l \cdot \| \boldsymbol{\xi} \|^\alpha,
\label{eq:mode_weighting}
\end{equation}
where $\gamma_l$ is a learnable scalar parameter at layer $l$ that controls the strength of the weighting;
$\alpha$ is a hyperparameter ($0.7$ in our experiments or optionally a learnable parameter) that determines how the weight scales with the frequency magnitude $\| \boldsymbol{\xi} \|$.
$\mathbf{w}(\boldsymbol{\xi})$ assigns higher weights to higher frequencies when $\alpha > 0$, thus emphasizing high-frequency components. This yields an updated $\mathcal{K}_l$:
\begin{equation}
\mathcal{K}_l \mathbf{v}_l(\mathbf{r}) = \mathcal{F}^{-1} \left( \mathbf{w}_l(\boldsymbol{\xi}) \cdot \mathbf{P}_l(\boldsymbol{\xi}) \cdot \mathcal{F}[\mathbf{v}_l](\boldsymbol{\xi}) \right).
\label{eq:adaptive_integral_operator}
\end{equation}

\subsection{\NewGatedFusionLong}
While \NewFNOShort excels at capturing global dependencies through spectral convolutions, it may not fully exploit local interactions critical for detailed image reconstruction. We incorporate \SpatialNOShort to complement \NewFNOShort by capturing local dependencies. Working in tandem, they learn mappings from the low-resolution input function to the high-resolution output function. \NewGatedFusionLong optimally combines the complementary features from both operators, adaptively balancing the contributions of each to a fused feature map, which is then fed to a projection layer.

Efficient implementation of the kernel integral using the Galerkin-type attention mechanism \cite{Galerkin} has been explored in the neural operator applied to SR tasks \cite{SR-SRNO, SR-HiNOTE}. Our \SpatialNOShort is composed of bicubic interpolation, Galerkin attention, and nonlinearity, sharing an encoder with \NewFNOShort. \SpatialNOShort models local interactions in the spatial domain, focusing on the most relevant spatial regions during the convolution process. Given the complementary role of \SpatialNOShort to \NewFNOShort, we simplify its structure to improve runtime. 

While previous works have applied gating mechanisms in different contexts, our approach differs significantly. Zheng et al. \cite{Gated-Fusion-OG} use gating within recurrent CRF networks primarily for semantic segmentation, controlling the information flow for boundary refinement rather than fusing feature maps with distinct representations. Hu et al. \cite{Gated-Fusion-Squeeze} introduced channel-wise gating in Squeeze-and-Excitation (SE) blocks, focusing on adaptively recalibrating feature channels within a single network stream. In contrast, our \NewGatedFusionLong applies spatial gating to integrate global dependencies captured by \NewFNOShort and local information from \SpatialNOShort. This mechanism adaptively combines both operators' feature maps, enhancing high-resolution image reconstruction by balancing global and local contributions at each spatial location. 

Let \( \mathbf{v}_{\text{\NewFNOShort}} \in \mathbb{R}^{B \times H \times W \times C} \) and \( \mathbf{v}_{\text{\SpatialNOShort}} \in \mathbb{R}^{B \times H \times W \times C} \) denote the feature maps obtained from \NewFNOShort and \SpatialNOShort, respectively, where \( B \) is the batch size, \( H \) and \( W \) are the height and width of the feature maps, and \( C \) is the number of channels. We first concatenate the feature maps along the channel dimension and pass them through a convolutional layer followed by a sigmoid activation to produce a gating map \( \mathbf{G} \in \mathbb{R}^{B \times H \times W \times 1} \):
\begin{equation}
\mathbf{G} = \sigma\left( \text{Conv}_{1 \times 1}\left( \left[ \mathbf{v}_{\text{\NewFNOShort}}, \mathbf{v}_{\text{\SpatialNOShort}} \right] \right) \right),
\label{eq:gating_map}
\end{equation}
where \( \left[ \cdot, \cdot \right] \) denotes concatenation along the channel dimension;
\( \text{Conv}_{1 \times 1} \) is a \( 1 \times 1 \) convolutional layer that reduces the concatenated features to a single-channel gating map;
\( \sigma(\cdot) \) is the sigmoid activation function applied element-wise.

The fused feature map \( \mathbf{v}_{\text{fused}} \in \mathbb{R}^{B \times H \times W \times C} \) is the element-wise weighted sum of the two feature maps:
\begin{equation}
\mathbf{v}_{\text{fused}} = \mathbf{G} \odot \mathbf{v}_{\text{\NewFNOShort}} + \left( 1 - \mathbf{G} \right) \odot \mathbf{v}_{\text{\SpatialNOShort}},
\label{eq:fused_features}
\end{equation}
where \( \odot \) denotes element-wise multiplication, and subtraction is performed element-wise. The gating map \( \mathbf{G} \) is broadcast across the channel dimension to match the dimensions of the feature maps.

\NewGatedFusionLong brings two advantages compared to a naive concatenation strategy:
(i) Captures complementary Information: \NewFNOShort models global dependencies through spectral convolutions, effectively modeling long-range interactions and overall structure. In contrast, \SpatialNOShort excels at capturing local dependencies and fine-grained details via attention mechanisms.
(ii) Balances contributions dynamically: \NewGatedFusionLong elicits the importance of each feature map at each spatial location, dynamically balancing global and local information.

\subsection{Forward Diffusion Process and Noise Schedule}
\noindent \textbf{Motivation}. NOs are well-suited for SR tasks due to their inherent resolution invariance and their ability to model global dependencies efficiently. Diffusion models can iteratively refine a low-resolution image to a high-resolution one, capturing the complex conditional distribution of high-resolution images given low-resolution inputs.

\MethodShort leverages the strengths of both frameworks. \NewFNOShort is a powerful mechanism for handling arbitrary resolutions and capturing high-frequency details, while the diffusion process iteratively improves reconstruction output.

In our framework, the forward diffusion process models the degradation of HR images to LR images, which in our case is primarily due to downscaling. To incorporate this degradation into the diffusion model framework, we define a forward process that simulates the downscaling effect over continuous time \( t \in [0, T] \). At, the image $\mathbf{x}_T$ closely resemble the observed LR image $\mathbf{x}_\text{LR}$ after significant degradation. We adopt a modified variance-preserving (VP) stochastic differential equation (SDE):
\begin{equation}
d\mathbf{x}_t = -\frac{1}{2} \beta(t) \left( \mathbf{x}_t - \mathbf{D}\mathbf{x}_t \right) dt + \sqrt{\beta(t)} d\mathbf{w},
\label{eq:forward_sde_modified}
\end{equation}
where \( \beta(t) \) is the noise schedule;
\( \mathbf{D} \) is the downsampling operator that reduces the resolution of the image;
\( d\mathbf{w} \) is the standard Wiener process.

In this formulation, the term \( \mathbf{x} - \mathbf{D}\mathbf{x} \) quantifies the high-frequency details lost during downscaling. The drift term \( -\frac{1}{2} \beta(t) \left( \mathbf{x} - \mathbf{D}\mathbf{x} \right) dt \) models the gradual removal of these details, while the diffusion term \( \sqrt{\beta(t)} d\mathbf{w} \) adds Gaussian noise to simulate further degradation.

\noindent\textbf{Noise Schedule} \( \beta(t) \). We define the noise schedule \( \beta(t) \) as a simple and effective linear function increasing over the time interval \( [0, T] \):
\begin{equation}
\beta(t) = \beta_{\text{min}} + (\beta_{\text{max}} - \beta_{\text{min}}) \cdot \frac{t}{T},
\label{eq:beta_schedule_linear}
\end{equation}
where \( \beta_{\text{min}} {=} 0.1 \) and \( \beta_{\text{max}} {=} 20 \).
This linear schedule ensures a gradual increase in the degradation strength from minimal degradation at \( t {=} 0 \) to maximum degradation at \( t {=} T \).

\noindent\textbf{Relation to Image Degradation}. At each time \( t \), the image \( \mathbf{x}_t\) progressively loses high-frequency details due to the drift toward \( \mathbf{D}\mathbf{x}_t \), the downscaled version of the image. The added Gaussian noise further simulates the information loss inherent in downscaling.
At \( t = T \), the image \( \mathbf{x}_T \) approximates the observed low-resolution image \( \mathbf{x}_{\text{LR}} \). The reverse diffusion process then aims to recover the high-resolution image \( \mathbf{x}_0 \) (i.e.\( \mathbf{x}_\text{HR} \) ) from \( \mathbf{x}_T \) by reversing the degradation.

\noindent\textbf{Choice of} \( \beta(t) \). The linear noise schedule is chosen for its simplicity and effectiveness. It provides a straightforward way to control the rate of degradation over time. Parameters \( \beta_{\text{min}} \) and \( \beta_{\text{max}} \) are selected to balance the trade-off between sufficient degradation (to simulate downscaling) and numerical stability of the diffusion process.

\noindent\textbf{Downsampling Operator} \( \mathbf{D} \). The operator \( \mathbf{D} \) is defined to reduce the spatial dimensions of the image by the desired scaling factor. We use bicubic downsampling.

\subsection{\NewSolverLong}
The standard reverse diffusion process is stochastic and requires a large number of sampling steps, making it computationally expensive. To accelerate inference, we reformulate the reverse diffusion as a deterministic Ordinary Differential Equation (ODE), allowing us to use advanced ODE solvers for faster sampling. The ODE solver integrates the reverse diffusion process, and its output is the super-resolved image.
The reverse diffusion process can be described by a Stochastic Differential Equation (SDE) \cite{SDE}:
\begin{equation}
d\mathbf{x} = \left[ f(\mathbf{x}, t) - g(t)^2 \nabla_{\mathbf{x}} \log p_t(\mathbf{x}) \right] dt + g(t) d\bar{\mathbf{w}}
\label{eq:reverse_diffusion_sde}
\end{equation}
where $\mathbf{x}$ is the data;
$t$ is the time variable;
$f(\mathbf{x}, t)$ and $g(t)$ are drift and diffusion coefficients;
$\nabla_{\mathbf{x}} \log p_t(\mathbf{x})$ is the score function;
$\bar{\mathbf{w}}$ is the reverse-time Wiener process. By removing the stochastic term, we obtain the probability flow ODE, which deterministically transports the data from the noise distribution to the data distribution.

Our \NewSolverShort ODE solver comprises three key components:

\noindent \textbf{1. Adaptive Time Step Selection Using a Learned Function}.
Optimizing the allocation of time steps based on data characteristics has been explored in previous works \cite{Adaptive-AutoDiffusion, Adaptive-AlignYourSteps}. We discretize the time interval $[0, T]$ into $N$ non-uniform time steps $\{ t_i \}_{i=0}^N$, where $t_0 = 0$ and $t_N = T$. We introduce a learned function $\phi_\psi(t)$ as a weighted sum of polynomial basis functions, where the weight is parameterized by a set of learnable coefficients $\psi = \{\psi_1, \psi_2, \dots, \psi_K\}$, which adaptively determines the distribution of time steps based on the data characteristics.

\noindent \textit{Parameterization of $\phi_\psi(t)$}.
We define $\phi_\psi(t)$ as a normalized weighted sum of $K$ predefined monotonically increasing basis functions \( \{\phi_k(t)\}_{k=1}^K \):
\begin{equation}
\phi_\psi(t) = \frac{\sum_{k=1}^K \psi_k \phi_k(t)}{\sum_{k=1}^K \psi_k \phi_k(T)}, \quad \psi_k = \exp(\omega_k),
\label{eq:phi_parameterization}
\end{equation}
where each basis function \( \phi_k(t) = t^k \) for \( k = 1, 2, \dots, K \) is polynomial. \( \omega_k \) are unconstrained learnable parameters that ensure \( \psi_k \geq 0 \) through the exponential mapping. We set $K=3$ to balance model flexibility with computational efficiency. This setup allows $\phi_\psi(t)$ to capture nonlinear time-step distributions without excessive complexity.

\noindent \textit{Selection of Time Steps}.
Using the learned function $\phi_\psi(t)$, we map uniformly spaced normalized values $s_i = \frac{i}{N}$ to non-uniform time steps $t_i$:
\begin{equation}
t_i = \phi_\psi^{-1}\left( s_i \right) = \phi_\psi^{-1}\left( \frac{i}{N} \right), \quad i = 0, 1, \dots, N
\label{eq:adaptive_time_steps_learned}
\end{equation}

Since $\phi_\psi(t)$ is monotonically increasing, its inverse function $\phi_\psi^{-1}(s)$ exists and can be efficiently computed.

\noindent \textbf{2. Neural Operator Score Network}.
The score function, representing the gradient of the log probability density \(\log p_t(\mathbf{x})\), is approximated using a neural network \(s_\theta(\mathbf{x}, t)\) parameterized by \(\theta\):
\begin{equation}
\nabla_{\mathbf{x}} \log p_t(\mathbf{x}) \approx s_\theta(\mathbf{x}, t).
\label{eq:score_function_approx}
\end{equation}
In our architecture, \(s_\theta\) consists of:
(i) An encoder that extracts features from \(\mathbf{x}_{\text{LR}}\));
(ii) \NewFNOShort for capturing global dependencies and high-frequency details;
(iii) \SpatialNOShort for modeling local dependencies and fine-grained structures;
(iv) \NewGatedFusionLong to dynamically combine features;
(v) Time embedding $\mathbf{e}(t)$ incorporating the time variable \( t \) into our neural network \( s_\theta(\mathbf{x}, t) \) using sinusoidal positional embeddings \cite{Attention, DDPM}, concatenating it and encoded features along the channel dimension.

\noindent \textbf{3. Efficient Solver}.
We solve the reverse-time stochastic differential equation (SDE) of the diffusion process, transformed into an ODE:
\begin{equation}
\frac{d\mathbf{x}}{dt} = f(\mathbf{x}, t) - \frac{1}{2} g(t)^2 \nabla_{\mathbf{x}} \log p_t(\mathbf{x}),
\label{eq:reverse_sde}
\end{equation}
where \(\mathbf{x} \in \mathbb{R}^d\) is the image estimate at time \(t\) in the reverse diffusion process;
\(f(\mathbf{x}, t)\) and \(g(t)\) are coefficients derived from the forward diffusion process.

For the Variance Preserving (VP) SDE commonly used in diffusion models, the coefficients are defined as:
\begin{equation}
f(\mathbf{x}, t) = -\frac{1}{2} \beta(t) \mathbf{x}, \quad g(t) = \sqrt{\beta(t)},
\label{eq:vp_sde_coefficients}
\end{equation}
where \(\beta(t)\) is a predefined noise schedule specific to the diffusion process and is consistent with our DiffFNO.
By substituting the score function approximation from Eq.~\eqref{eq:score_function_approx}, we define the approximate drift function:
\begin{equation}
f_\theta(\mathbf{x}, t) = f(\mathbf{x}, t) - \frac{1}{2} g(t)^2 s_\theta(\mathbf{x}, t).
\label{eq:approximate_drift}
\end{equation}

The adaptive time steps \(\{ t_i \}_{i=0}^N\) discretize the ODE. We apply the Runge-Kutta 4th-order (RK4) method, as it balances computational cost and accuracy, requiring fewer steps than lower-order methods while retaining precision.

The benefits of \NewSolverShort are threefold:
(i) Deterministic Sampling: It consistently produces the same results for identical inputs, improving reproducibility. 
(ii) Reduced Computation: Fewer sampling steps significantly decrease inference time. 
(iii) High-Quality Reconstruction: It maintains high-quality reconstruction by efficiently allocating computational resources.
\section{Experiments}

\newcommand{\imgheight}{1.8cm} 
\newcommand{\imgspacing}{0.2cm} 
\newcommand{\firstimgheight}{4.2cm} 

\begin{figure*}[t]
  \centering
  \begin{tabular}{@{}c@{\hspace{\imgspacing}}c@{}}
    
    \begin{tabular}{c}
      \includegraphics[height=\firstimgheight]{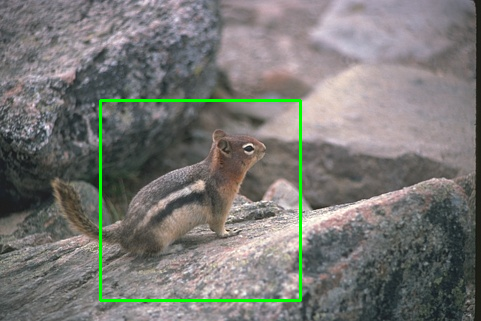} \\ 
      \small BSD100 \cite{BSD100}, $\times 12$
    \end{tabular}
    &
    
    \begin{tabular}{
      c@{\hskip \imgspacing} 
      c@{\hskip \imgspacing} 
      c@{\hskip \imgspacing} 
      c@{\hskip \imgspacing} 
      c}
      
      \includegraphics[height=\imgheight]{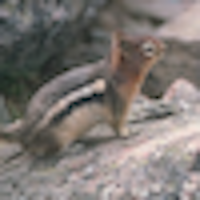} &
      \includegraphics[height=\imgheight]{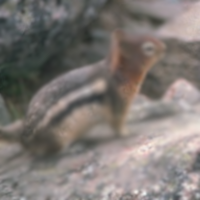} &
      \includegraphics[height=\imgheight]{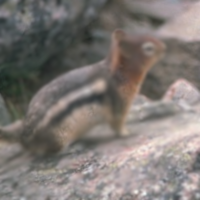} &
      \includegraphics[height=\imgheight]{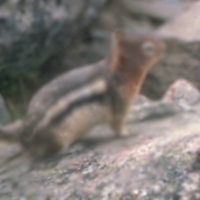} &
      \includegraphics[height=\imgheight]{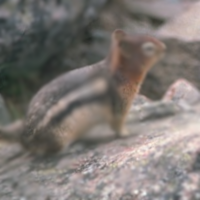} \\

      \small Bicubic & \small Meta-SR \cite{SR-Meta-SR} & 
      \small LTE \cite{SR-LTE} & \small LIIF \cite{SR-LIIF} & \small LIT \cite{SR-LIT} \\

      \includegraphics[height=\imgheight]{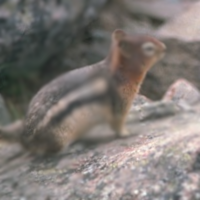} &
      \includegraphics[height=\imgheight]{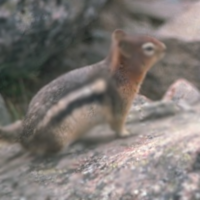} &
      \includegraphics[height=\imgheight]{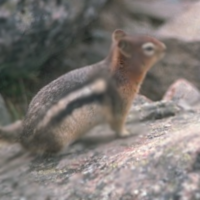} &
      \includegraphics[height=\imgheight]{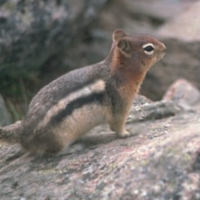} &
      \includegraphics[height=\imgheight]{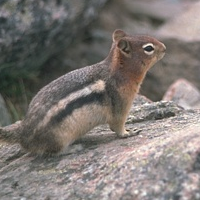} \\

      \small LMI \cite{SR-LMI} & \small SRNO \cite{SR-SRNO} & 
      \small HiNOTE \cite{SR-HiNOTE} & \small DiffFNO (ours) & \small GT \\
      
    \end{tabular}
    
  \end{tabular}
  
  \begin{tabular}{@{}c@{\hspace{\imgspacing}}c@{}}
    
    \begin{tabular}{c}
      \includegraphics[height=\firstimgheight]{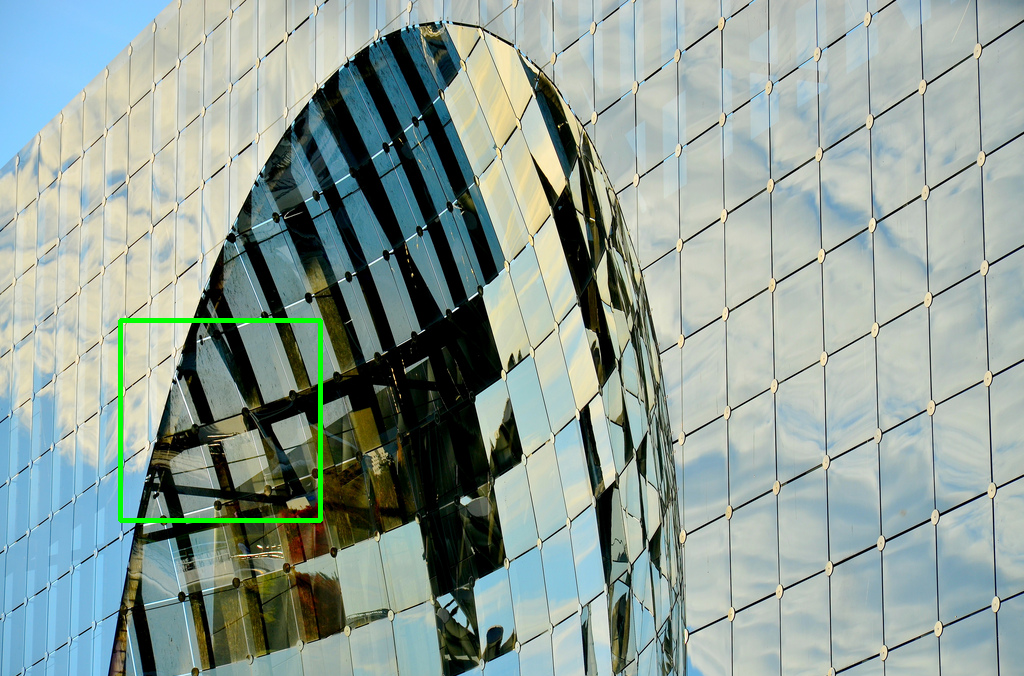} \\ 
      \small Urban100 \cite{URBAN100}, $\times 7.6$
    \end{tabular}
    &
    
    \begin{tabular}{
      c@{\hskip \imgspacing} 
      c@{\hskip \imgspacing} 
      c@{\hskip \imgspacing} 
      c@{\hskip \imgspacing} 
      c}
      
      \includegraphics[height=\imgheight]{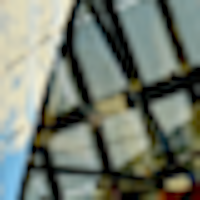} &
      \includegraphics[height=\imgheight]{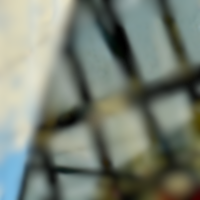} &
      \includegraphics[height=\imgheight]{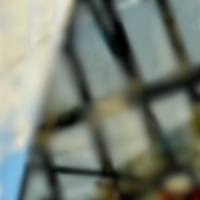} &
      \includegraphics[height=\imgheight]{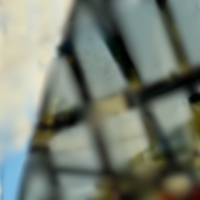} &
      \includegraphics[height=\imgheight]{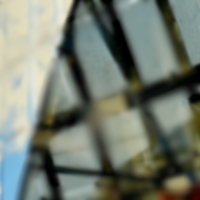} \\

      \small Bicubic & \small Meta-SR \cite{SR-Meta-SR} & 
      \small LTE \cite{SR-LTE} & \small LIIF \cite{SR-LIIF} & \small LIT \cite{SR-LIT} \\

      \includegraphics[height=\imgheight]{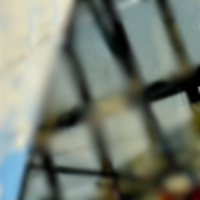} &
      \includegraphics[height=\imgheight]{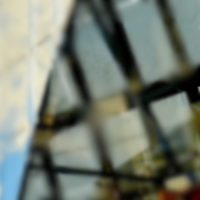} &
      \includegraphics[height=\imgheight]{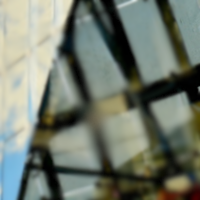} &
      \includegraphics[height=\imgheight]{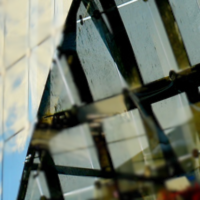} &
      \includegraphics[height=\imgheight]{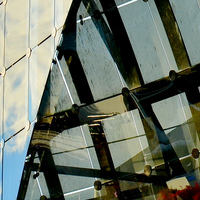} \\

      \small LMI \cite{SR-LMI} & \small SRNO \cite{SR-SRNO} & 
      \small HiNOTE \cite{SR-HiNOTE} & \small DiffFNO (ours) & \small GT \\
      
    \end{tabular}
    
  \end{tabular}
  
  \caption{Qualitative comparison on integer and continuous super-resolution scales. The models use RDN \cite{RDN} as their encoder (except HiNOTE \cite{SR-HiNOTE}, has its own). In the HR image, the cropped patch is outlined in green.}
  \label{fig:qualitative_comparison}
    \vspace{-0.4cm}
\end{figure*}

\begin{table*}[htbp]
  \centering

  \resizebox{\textwidth}{!}{  
    \begin{tabular}{lcccccccccccc}
      \toprule
      \multirow{2}{*}{\textbf{Model}} & \multicolumn{2}{c}{\textbf{×2}} & \multicolumn{2}{c}{\textbf{×3}} & \multicolumn{2}{c}{\textbf{×4}} & \multicolumn{2}{c}{\textbf{×6}} & \multicolumn{2}{c}{\textbf{×8}} & \multicolumn{2}{c}{\textbf{×12}} \\
      \cmidrule(lr){2-3} \cmidrule(lr){4-5} \cmidrule(lr){6-7} \cmidrule(lr){8-9} \cmidrule(lr){10-11} \cmidrule(lr){12-13}
      & \textbf{PSNR} & \textbf{SSIM} & \textbf{PSNR} & \textbf{SSIM} & \textbf{PSNR} & \textbf{SSIM} & \textbf{PSNR} & \textbf{SSIM} & \textbf{PSNR} & \textbf{SSIM} & \textbf{PSNR} & \textbf{SSIM} \\
      \midrule
      EDSR-MetaSR \cite{SR-Meta-SR} & 33.32 & 0.913 & 30.10 & 0.800 & 28.23 & 0.830 & 26.10 & 0.792 & 24.77 & 0.742 & 23.95 & 0.720 \\
      EDSR-LTE \cite{SR-LTE} & 33.83 & 0.921 & 30.50 & 0.880 & 28.79 & 0.852 & 26.55 & 0.800 & 25.05 & 0.760 & 24.20 & 0.736 \\
      EDSR-LIIF \cite{SR-LIIF} & 34.36 & 0.925 & 30.94 & 0.885 & 29.31 & 0.855 & 27.02 & 0.814 & 25.44 & 0.771 & 24.32 & 0.743 \\
      EDSR-LIT \cite{SR-LIT} & 34.81 & 0.928 & 31.39 & 0.890 & 29.70 & 0.860 & 27.44 & 0.815 & 25.78 & 0.775 & 24.69 & 0.745 \\
      EDSR-LMI \cite{SR-LMI} & 35.40 & 0.930 & 31.88 & 0.895 & 30.40 & 0.865 & 27.95 & 0.820 & 26.16 & 0.780 & 25.56 & 0.750 \\
      EDSR-SRNO \cite{SR-SRNO} & 34.85 & 0.928 & 31.45 & 0.890 & 30.05 & 0.863 & 27.36 & 0.810 & 26.00 & 0.772 & 25.91 & 0.760 \\
      \textbf{EDSR-DiffFNO (Ours)} & \textbf{35.72} & \textbf{0.932} & \textbf{32.50} & \textbf{0.905} & \textbf{30.88} & \textbf{0.870} & \textbf{28.29} & \textbf{0.830} & \textbf{26.78} & \textbf{0.790} & \textbf{26.48} & \textbf{0.775} \\
      \midrule
      HiNOTE$^\dagger$ \cite{SR-HiNOTE} & 35.29 & 0.931 & 31.90 & 0.895 & 30.46 & 0.842 & 27.83 & 0.799 & 26.41 & 0.772 & 26.23 & 0.732 \\
      \midrule
      RDN-MetaSR \cite{SR-Meta-SR} & 33.50 & 0.920 & 30.32 & 0.893 & 28.41 & 0.861 & 26.29 & 0.810 & 24.90 & 0.780 & 24.01 & 0.790 \\
      RDN-LTE \cite{SR-LTE} & 33.98 & 0.922 & 30.65 & 0.882 & 28.94 & 0.852 & 26.70 & 0.802 & 25.20 & 0.762 & 24.35 & 0.732 \\
      RDN-LIIF \cite{SR-LIIF} & 34.51 & 0.927 & 31.09 & 0.887 & 29.46 & 0.857 & 27.17 & 0.812 & 25.59 & 0.772 & 24.47 & 0.742 \\
      RDN-LIT \cite{SR-LIT} & 34.96 & 0.930 & 31.54 & 0.892 & 29.85 & 0.862 & 27.59 & 0.817 & 25.93 & 0.777 & 24.84 & 0.747 \\
      RDN-LMI \cite{SR-LMI} & 35.55 & 0.932 & 32.03 & 0.897 & 30.55 & 0.867 & 28.10 & 0.822 & 26.31 & 0.782 & 25.71 & 0.752 \\
      RDN-SRNO \cite{SR-SRNO} & 35.00 & 0.930 & 31.60 & 0.892 & 30.20 & 0.862 & 27.51 & 0.812 & 26.15 & 0.772 & 26.06 & 0.762 \\
      \textbf{RDN-DiffFNO (Ours)}  & \textbf{35.87} & \textbf{0.934} & \textbf{32.65} & \textbf{0.902} & \textbf{31.03} & \textbf{0.872} & \textbf{28.44} & \textbf{0.832} & \textbf{26.93} & \textbf{0.792} & \textbf{26.63} & \textbf{0.777} \\
      \bottomrule
    \end{tabular}
  }
  \caption{PSNR/SSIM comparison on the DIV2K \cite{DIV2K} validation set using EDSR \cite{EDSR} and RDN \cite{RDN} encoders. HiNOTE \cite{SR-HiNOTE} uses its own.}
  \label{tab:comparison-quantitative}
  \vspace{-0.4cm}
\end{table*}

\begin{table*}[t]
  \centering

  \setlength{\tabcolsep}{2pt}
  
  \resizebox{\textwidth}{!}{ 
    \begin{tabular}{lcccccccccccccccccccc}
      \toprule
      \multirow{2}{*}{\textbf{Model}} & \multicolumn{5}{c}{\textbf{Set5}} & \multicolumn{5}{c}{\textbf{Set14}} & \multicolumn{5}{c}{\textbf{BSD100}} & \multicolumn{5}{c}{\textbf{Urban100}} \\
      \cmidrule(lr){2-6} \cmidrule(lr){7-11} \cmidrule(lr){12-16} \cmidrule(lr){17-21}
       & \textbf{$\times$2} & \textbf{$\times$3} & \textbf{$\times$4} & \textbf{$\times$6} & \textbf{$\times$8} & \textbf{$\times$2} & \textbf{$\times$3} & \textbf{$\times$4} & \textbf{$\times$6} & \textbf{$\times$8} & \textbf{$\times$2} & \textbf{$\times$3} & \textbf{$\times$4} & \textbf{$\times$6} & \textbf{$\times$8} & \textbf{$\times$2} & \textbf{$\times$3} & \textbf{$\times$4} & \textbf{$\times$6} & \textbf{$\times$8} \\
      \midrule
      MetaSR \cite{SR-Meta-SR}& 37.50 & 34.05 & 31.52 & 28.23 & 26.02 & 33.51 & 30.03 & 28.02 & 25.53 & 24.02 & 31.02 & 28.05 & 26.52 & 24.82 & 23.52 & 32.02 & 28.03 & 25.82 & 23.52 & 22.03 \\
      LIIF \cite{SR-LIIF}& 38.02 & 34.42 & 32.04 & 28.57 & 26.25 & 34.03 & 30.43 & 28.43 & 25.84 & 24.33 & 31.52 & 28.55 & 27.03 & 25.03 & 23.83 & 32.52 & 28.53 & 26.03 & 23.83 & 22.33 \\
      LTE \cite{SR-LTE}& 38.21 & 34.63 & 32.25 & 28.76 & 26.44 & 34.22 & 30.65 & 28.64 & 26.05 & 24.52 & 31.71 & 28.73 & 27.23 & 25.23 & 24.03 & 32.72 & 28.75 & 26.23 & 24.03 & 22.53 \\
      SRNO \cite{SR-SRNO}& 38.32 & 34.84 & 32.69 & 29.38 & 27.28 & 34.27 & 30.71 & 28.97 & 26.76 & 25.26 & 32.43 & 29.37 & 27.83 & 26.04 & 24.99 & 33.33 & 29.14 & 26.98 & 24.43 & 23.02 \\
      LIT \cite{SR-LIT}& 38.53 & 35.02 & 32.82 & 29.51 & 27.42 & 34.44 & 30.83 & 29.03 & 26.82 & 25.33 & 32.52 & 29.51 & 27.92 & 26.12 & 25.01 & 33.42 & 29.22 & 27.02 & 24.52 & 23.12 \\
      LMI \cite{SR-LMI}& 38.72 & 35.14 & 32.95 & 29.63 & 27.55 & 34.63 & 31.02 & 29.24 & 27.05 & 25.55 & 32.72 & 29.74 & 28.04 & 26.25 & 25.14 & 33.62 & 29.44 & 27.24 & 24.63 & 23.23 \\
      HiNOTE \cite{SR-HiNOTE} & 39.01 & 35.22 & 33.08 & 29.85 & 27.74 & 35.02 & 31.25 & 29.55 & 27.35 & 25.85 & 33.02 & 30.05 & 28.15 & 26.35 & 25.25 & 34.03 & 29.83 & 27.55 & 24.73 & 23.34 \\
      \textbf{DiffFNO (Ours)} & \textbf{39.72} & \textbf{35.30} & \textbf{33.16} & \textbf{30.23} & \textbf{27.93} & \textbf{36.01} & \textbf{31.54} & \textbf{30.22} & \textbf{27.58} & \textbf{26.02} & \textbf{33.56} & \textbf{30.24} & \textbf{28.21} & \textbf{26.45} & \textbf{25.30} & \textbf{34.19} & \textbf{29.99} & \textbf{27.74} & \textbf{24.80} & \textbf{23.35} \\
      \bottomrule
    \end{tabular}
  }
  \caption{PSNR comparison on four benchmark datasets: Set5 \cite{SET5}, Set14 \cite{SET14}, BSD100 \cite{BSD100}, and Urban100 \cite{URBAN100}. All models use RDN \cite{RDN} as their encoder, besides HiNOTE \cite{SR-HiNOTE} which has its own.}
  \label{tab:comparison-benchmarks}
    \vspace{-0.4cm}
\end{table*}

\noindent\textbf{Datasets and Evaluation Metrics.}
We use the DIV2K \cite{DIV2K} dataset for training. For evaluation, we use the DIV2K validation set and four standard datasets: Set5 \cite{SET5}, Set14 \cite{SET14}, BSD100 \cite{BSD100}, and Urban100 \cite{URBAN100}.
    
We evaluate our model on upscaling factors of $\times$2, $\times$3, $\times$4, $\times$6, $\times$8, and $\times$12. Notably, scales $\times$6, $\times$8, and $\times$12 are outside the training distribution, as the training scales are uniformly sampled from $\times$1 to $\times$4. This setup assesses our model's ability to generalize to arbitrary scales. We use Peak Signal-to-Noise Ratio (PSNR) and Structural Similarity Index Measure (SSIM) as our evaluation metrics.
    
\noindent\textbf{Quantitative Results.}
Building on the quantitative gains of our model, we also present qualitative results to illustrate the visual improvements achieved. We compare our proposed \MethodShort model with several SOTA arbitrary-scale SR methods, including Meta-SR \cite{SR-Meta-SR}, LIIF \cite{SR-LIIF}, LTE \cite{SR-LTE}, SRNO \cite{SR-SRNO}, LIT \cite{SR-LIT}, LMI \cite{SR-LMI}, and HiNOTE \cite{SR-HiNOTE}. All models are trained on the DIV2K dataset with identical settings to ensure a fair comparison in Tables \ref{tab:comparison-quantitative} and \ref{tab:comparison-benchmarks}.

Among the compared methods, Meta-SR performs adequately at lower scales but struggles at higher scaling factors due to its generalized approach that lacks specialized mechanisms for fine detail capture. LIIF and LTE improve upon Meta-SR by using local implicit functions and frequency-based estimations, respectively, which enhance high-frequency texture representation. However, they still face limitations in capturing non-periodic textures and high-frequency details, resulting in blurred textures at larger scales. LIT and LMI further advance performance by integrating attention mechanisms and MLP-mixer architectures, effectively preserving high-frequency textures and handling diverse scales, but they may not generalize well across datasets with varying distributions. SRNO and HiNOTE employ neural operator frameworks with attention mechanisms and frequency-aware loss priors to better capture global spatial properties and enhance high-frequency detail reconstruction. We observed mixed results between SRNO and HiNOTE: at certain scaling factors, one outperforms the other, indicating their varying strengths at different resolutions. Overall, their neural operator foundation improves the handling of arbitrary scaling but may increase computational demands.

\begin{table*}[htbp]
  \centering
  \resizebox{\textwidth}{!}{
    \begin{tabular}{@{}l@{\hskip 0.3cm} c c c c c c c c c c c c@{\hskip 0.3cm} c@{\hskip 0.3cm} c@{}}
      \toprule
      \multirow{2}{*}{\textbf{Model}} & \multicolumn{2}{c}{\textbf{$\times$2}} & \multicolumn{2}{c}{\textbf{$\times$3}} & \multicolumn{2}{c}{\textbf{$\times$4}} & \multicolumn{2}{c}{\textbf{$\times$6}} & \multicolumn{2}{c}{\textbf{$\times$8}} & \multicolumn{2}{c}{\textbf{$\times$12}} & \multirow{2}{*}{\textbf{Inference}} & \multirow{2}{*}{\textbf{Steps}} \\
      \cmidrule(lr){2-3} \cmidrule(lr){4-5} \cmidrule(lr){6-7} \cmidrule(lr){8-9} \cmidrule(lr){10-11} \cmidrule(lr){12-13}
       & \textbf{PSNR} & \textbf{SSIM} & \textbf{PSNR} & \textbf{SSIM} & \textbf{PSNR} & \textbf{SSIM} & \textbf{PSNR} & \textbf{SSIM} & \textbf{PSNR} & \textbf{SSIM} & \textbf{PSNR} & \textbf{SSIM} & & \\
      \midrule
      SRNO \cite{SR-SRNO} & 33.81 & 0.920 & 30.53 & 0.880 & 28.74 & 0.850 & 26.59 & 0.800 & 25.10 & 0.760 & 24.18 & 0.730 & 147 & - \\
      FNO \cite{FNO} & 34.36 & 0.925 & 30.94 & 0.885 & 29.31 & 0.855 & 27.02 & 0.810 & 25.44 & 0.770 & 24.32 & 0.740 & \textbf{85} & - \\
      \midrule
      WFNO & 34.81 & 0.928 & 31.39 & 0.888 & 29.70 & 0.858 & 27.44 & 0.815 & 25.78 & 0.775 & 24.69 & 0.745 & 97 & - \\
      WFNO-AttnNO & 35.40 & 0.930 & 31.88 & 0.892 & 30.40 & 0.862 & 27.95 & 0.820 & 26.16 & 0.780 & 25.56 & 0.750 & 139 & 1000 \\
      DiffFNO(-w, -a, -s) & 34.85 & 0.928 & 31.45 & 0.890 & 30.05 & 0.860 & 27.36 & 0.815 & 26.00 & 0.775 & 25.91 & 0.760 & 204 & 1000 \\
      DiffFNO(-a, -s) & 35.29 & 0.930 & 31.90 & 0.893 & 30.46 & 0.863 & 27.83 & 0.820 & 26.41 & 0.780 & 26.23 & 0.765 & 231 & 1000 \\
      DiffFNO(-s) & 35.70 & \textbf{0.932} & 32.48 & 0.896 & 30.85 & 0.866 & 28.26 & 0.825 & 26.75 & 0.785 & 26.45 & 0.770 & 266 & 1000 \\
      \textbf{\MethodShort} & \textbf{35.72} & \textbf{0.932} & \textbf{32.50} & \textbf{0.900} & \textbf{30.88} & \textbf{0.870} & \textbf{28.29} & \textbf{0.830} & \textbf{26.78} & \textbf{0.790} & \textbf{26.48} & \textbf{0.775} & 141 & \textbf{30} \\
      \bottomrule
    \end{tabular}
  }
  \caption{Ablation study of variants of \MethodShort on the DIV2K \cite{DIV2K} validation set. All use EDSR-baseline \cite{EDSR} backbone as their encoder. Inference times are measured in milliseconds (ms). WFNO-AttnNO has \NewGatedFusionLong.}
  \label{tab:ablation_study}
    \vspace{-0.4cm}
\end{table*}

Our \MethodShort model consistently achieves the highest PSNR and SSIM scores across all scaling factors and datasets. The performance gap widens at larger scaling factors ($\times$8 and $\times$12), demonstrating a superior generalization to the out-of-distribution scales. The improvements are more pronounced on complex datasets like Urban100, which contain intricate textures and structures. By combining \NewFNOShort and \SpatialNOShort features through the \NewGatedFusionLong, which adaptively balances global and local features, \MethodShort synthesizes global and local dependencies. The \NewSolverShort ODE solver efficiently refines high-resolution images, further enhancing quality. This combination addresses the limitations of prior models, such as spectral bias and insufficient high-frequency detail capture.

\noindent\textbf{Qualitative Results.}
Fig. \ref{fig:qualitative_comparison} compares arbitrary-scale SR methods on a BSD100 image (scaling factor of $\times$12) with fine-grained details like animal fur and rock textures and an Urban100 (continuous scaling factor of $\times$7.6) image featuring large structures and fine local details such as reflections on the grass. SRNO and HiNOTE capture multiscale details effectively, from the animal’s body to tiny gaps between glass panels. However, \MethodShort reconstructs crisper edges with fewer artifacts, enhancing texture in animal fur pattern and reflections. \NewFNOShort captures large-scale patterns, while \SpatialNOShort and \NewGatedFusionLong preserve intricate textures. This multiscale approach followed by a diffusion process enhanced by the \NewSolverShort’s ODE solver further reduces visual artifacts.

\noindent\textbf{Ablation Studies.}
Extensive ablation studies validate the effectiveness and complementary nature of new components in \MethodShort. Table \ref{tab:ablation_study} reports the PSNR results on the DIV2K validation set for different model variants with scaling factors from $\times$2 to $\times$12. \textbf{-w} denotes leaving out the \NewFNOFeatureLong (yielding the default FNO \cite{FNO}). \textbf{-a} denotes omitting \SpatialNOShort. \textbf{-s} denotes the removal of \NewSolverShort ODE solver. We also measure inference time by averaging over $100$ runs, and report inference steps. We establish a baseline with SRNO, whose architecture is the most similar to our \MethodShort aamong the methods covered in our study. Overall, we observe notable improvements with the addition of model components. The complete \MethodShort achieves the highest PSNR and SSIM values across all upscaling factors. 

\noindent \textbf{Effect of \NewFNOFeatureLong}. Incorporating \NewFNOFeatureLong into \NewFNOShort boosted performance compared to the default FNO \cite{FNO}, at the cost of a slightly increased number of parameters and inference time.

\noindent \textbf{Effect of \NewGatedFusionLong and \SpatialNOShort}. The \NewGatedFusionLong introduces minimal computational overhead. In addition, extra computational cost incurred by \SpatialNOLong is effectively mitigated by running it in parallel with \NewFNOShort while employing a shared encoder.

\noindent \textbf{Effect of \NewSolverShort ODE Solver}: \NewSolverShort dramatically reduces the number of inference steps from $1,000$ to just $30$, which substantially improves the inference time while delivering competitive performance. As demonstrated in Tab. \ref{tab:ablation_study}, this acceleration not only preserves image quality but can even lead to slight improvements. In Fig. \ref{fig:teaser} (a), \MethodShort outperforms existing methods in both PSNR and inference time.

\section{Conclusion}
We propose \MethodLong (\MethodShort) for arbitrary-scale image super-resolution. \MethodShort is made of \NewFNOLong with a \NewFNOFeatureLong mechanism to emphasize high-frequency details. It is complemented by a \SpatialNOLong through a \NewGatedFusionLong that effectively adjusts the influence of global and local features. Image reconstruction is further refined by a diffusion process augmented with an \NewSolverLong ODE solver that dynamically allocates time steps, drastically cutting down inference time without compromising output quality. Experiments demonstrate \MethodShort's competitiveness in both reconstruction quality and inference time across various benchmarks, establishing a new state-of-the-art.
{
    \small
    \bibliographystyle{ieeenat_fullname}
    \bibliography{main}
}


\end{document}